\documentclass[10pt,journal,compsoc]{IEEEtran}
\ifCLASSOPTIONcompsoc
\usepackage[nocompress]{cite}
\else
\usepackage{cite}
\fi
\ifCLASSINFOpdf
\else
\fi

\usepackage{subfig}
\usepackage{array}
\usepackage{booktabs}
\usepackage[linesnumbered,ruled,vlined]{algorithm2e}
\usepackage{amsmath}

\DeclareMathOperator*{\argmax}{arg\,max}
\DeclareMathOperator*{\argmin}{arg\,min}

\SetKwInput{KwInput}{Input}        
\SetKwInput{KwOutput}{Output} 
\SetKwInOut{KwParameter}{Parameters}
\SetKwRepeat{Do}{do}{while}

\SetKwComment{tcc}{$\triangleright$\ }{}

\SetCommentSty{commentfont}

\newcolumntype{C}[1]{>{\centering}p{#1}}

\usepackage{tikz}
\usetikzlibrary{positioning}
\usetikzlibrary{patterns}
\usetikzlibrary{pgfplots.groupplots}
\tikzset{%
    block/.style = {rectangle, draw, 
        font=\scriptsize,
        text width=6.5em, text centered,
        rounded corners, minimum height=2em 
    },
}

\usepackage{pgfplots}
\usepgfplotslibrary{fillbetween}
\pgfplotsset{compat=newest, 
    tick label style={font=\scriptsize},
    label style={font=\scriptsize},
    legend style={font=\scriptsize}}

\newenvironment{customlegend}[1][]{%
    \begingroup
    \csname pgfplots@init@cleared@structures\endcsname
    \pgfplotsset{#1}%
}{%
    \csname pgfplots@createlegend\endcsname
    \endgroup
}%

\def\addlegendimage{\csname pgfplots@addlegendimage\endcsname}

\definecolor{brightBlue}{RGB}{68, 119, 170}
\definecolor{brightCyan}{RGB}{102, 204, 238}
\definecolor{brightGreen}{RGB}{34, 136, 51}
\definecolor{brightYellow}{RGB}{204, 187, 68}
\definecolor{brightRed}{RGB}{238, 102, 119}
\definecolor{brightPurple}{RGB}{170, 51, 119}
\definecolor{brightGrey}{RGB}{187, 187, 187}

\definecolor{vibrantBlue}{RGB}{0, 119, 187}
\definecolor{vibrantCyan}{RGB}{51, 187, 238}
\definecolor{vibrantTeal}{RGB}{0, 153, 136}
\definecolor{vibrantOrange}{RGB}{238, 119, 51}
\definecolor{vibrantRed}{RGB}{204, 51, 17}
\definecolor{vibrantMagenta}{RGB}{238, 51, 119}
\definecolor{vibrantGrey}{RGB}{100, 100, 100}

\tikzstyle{dotted}=                  [dash pattern=on \pgflinewidth off 2pt]
\tikzstyle{densely dotted}=          [dash pattern=on \pgflinewidth off 1pt]
\tikzstyle{loosely dotted}=          [dash pattern=on \pgflinewidth off 4pt]

\tikzstyle{dashed}=                  [dash pattern=on 3pt off 3pt]
\tikzstyle{densely dashed}=          [dash pattern=on 3pt off 2pt]
\tikzstyle{loosely dashed}=          [dash pattern=on 3pt off 6pt]

\tikzstyle{dashdotted}=              [dash pattern=on 3pt off 2pt on \pgflinewidth off 2pt]
\tikzstyle{densely dashdotted}=      [dash pattern=on 3pt off 1pt on \pgflinewidth off 1pt]
\tikzstyle{loosely dashdotted}=      [dash pattern=on 3pt off 4pt on \pgflinewidth off 4pt]

\tikzstyle{dash dot dot}=[dash pattern=on 3pt off 2pt on \pgflinewidth off 2pt on \pgflinewidth off 2pt]
\tikzstyle{densely dash dot dot}=[dash pattern=on 3pt off 1pt on \pgflinewidth off 1pt on \pgflinewidth off 1pt]
\tikzstyle{loosely dash dot dot}= [dash pattern=on 3pt off 4pt on \pgflinewidth off 4pt on \pgflinewidth off 4pt]

\tikzstyle{dash dash dot}=[dash pattern=on 3pt off 2pt on 3pt off 2pt on \pgflinewidth off 2pt]
\tikzstyle{densely dash dash dot}=[dash pattern=on 3pt off 1pt on 3pt off 1pt on \pgflinewidth off 1pt]
\tikzstyle{loosely dash dash dot}= [dash pattern=on 3pt off 4pt on 3pt off 4pt on \pgflinewidth off 4pt]

\tikzstyle{dash dash dot dot}=[dash pattern=on 3pt off 2pt on 3pt off 2pt on \pgflinewidth off 2pt on \pgflinewidth off 2pt]
\tikzstyle{densely dash dash dot dot}=[dash pattern=on 3pt off 1pt on 3pt off 1pt on \pgflinewidth off 1pt on \pgflinewidth off 1pt]
\tikzstyle{loosely dash dash dot dot}= [dash pattern=on 3pt off 4pt on 3pt off 4pt on \pgflinewidth off 4pt on \pgflinewidth off 4pt]

\pgfplotsset{
	cycle list/.define={vibrant}{
    	vibrantBlue, solid, every mark/.append style={solid, fill=vibrantBlue},mark=*\\
    	vibrantOrange, densely dashed, every mark/.append style={solid, fill=vibrantOrange}, mark=triangle*\\
    	vibrantTeal, densely dotted, every mark/.append style={solid, fill=vibrantTeal}, mark=square*\\
    	vibrantCyan, densely dashdotted, every mark/.append style={solid, fill=vibrantCyan}, mark=diamond*\\
        vibrantRed, densely dash dot dot, every mark/.append style={solid, fill=vibrantRed, scale=1.5}, mark=x\\
        vibrantGrey, densely dash dash dot, every mark/.append style={solid, fill=vibrantGrey, scale=1.5}, mark=star\\
        vibrantMagenta, densely dash dash dot dot, every mark/.append style={solid, fill=vibrantMagenta, scale=1.5}, mark=|\\
        },
    }

\usetikzlibrary{external}
\pgfkeys{/pgf/number format/.cd,1000 sep={\,}}

\begin{document}
    %
    \title{Tiny Machine Learning for Concept Drift}
    %
    %
    %
    %
    
    \author{Simone~Disabato,
        and~Manuel~Roveri,~\IEEEmembership{Senior~Member,~IEEE}
        \IEEEcompsocitemizethanks{\IEEEcompsocthanksitem S. Disabato and M. Roveri are with the Dipartimento di Elettronica, Informazione e Bioingegneria (DEIB), Politecnico di Milano, Milan, Italy.\protect\\
            E-mail: \{simone.disabato,manuel.roveri\}@polimi.it}
        \thanks{Manuscript received April 19, 2005; revised August 26, 2015.}}
    
    %
    %

    \markboth{Journal of \LaTeX\ Class Files,~Vol.~14, No.~8, August~2021}%
    {Shell \MakeLowercase{\textit{et al.}}: Bare Demo of IEEEtran.cls for Computer Society Journals}
    %



    \IEEEtitleabstractindextext{%
        \begin{abstract}
            Tiny Machine Learning (TML) is a new research area whose goal is to design machine and deep learning techniques able to operate in Embedded Systems and IoT units, hence satisfying the severe technological constraints on memory, computation, and energy characterizing these pervasive devices. Interestingly, the related literature mainly focused on reducing the computational and memory demand of the inference phase of machine and deep learning models. At the same time, the training is typically assumed to be carried out in Cloud or edge computing systems (due to the larger memory and computational requirements). This assumption results in TML solutions that might become obsolete when the process generating the data is affected by concept drift (e.g., due to periodicity or seasonality effect, faults or malfunctioning affecting sensors or actuators, or changes in the users' behavior), a common situation in real-world application scenarios.  For the first time in the literature, this paper introduces a Tiny Machine Learning for Concept Drift (TML-CD) solution based on deep learning feature extractors and a k-nearest neighbors classifier integrating a hybrid adaptation module able to deal with concept drift affecting the data-generating process. This adaptation module continuously updates (in a passive way) the knowledge base of TML-CD and, at the same time, employs a Change Detection Test to inspect for changes (in an active way) to quickly adapt to concept drift by removing the obsolete knowledge.
            Experimental results on both image and audio benchmarks show the effectiveness of the proposed solution, whilst the porting of TML-CD on three off-the-shelf micro-controller units shows the feasibility of what is proposed in real-world pervasive systems.
        \end{abstract}
        
}
      
    \maketitle

    \IEEEdisplaynontitleabstractindextext

    %
    \IEEEpeerreviewmaketitle
    
    %
    \IEEEraisesectionheading{
    \section{Introduction}
    \label{sec:introduction}
}
Internet-of-Things (IoT) and embedded systems are nowadays part of our everyday life in a wide range of application scenarios (e.g., automotive, medical devices, and smart cities, to name a few). In recent years, the scientific and technological trend about these pervasive devices is to move the processing (and in particular the intelligent processing) as close as possible to where data are generated. The reason is twofold. First, IoT units and embedded systems already operate pervasively in the environment processing large amounts of data acquired by the sensors. Second, machine and deep learning solutions processing these data directly on the pervasive devices are crucial to support real-time applications, prolong the system lifetime, and increase the Quality-of-Service. Nevertheless, machine and deep learning solutions are typically characterized by memory and computational demands that rarely match the constraints on memory, computation, and energy characterizing the IoT units and embedded systems~\cite{alippi2018moving,sanchez2020tinyml,tang2017enabling}. 

\emph{Tiny Machine Learning} (TML)~\cite{banbury2020benchmarking} is a relatively new research area aiming at filling this gap by designing ``tiny'' machine and deep learning solutions able to run on IoT units and embedded systems. Section~\ref{sct:literature} analyses the related literature, highlighting that most TML solutions focus on approximation, pruning, and quantization mechanisms to reduce memory and computational demand of machine and deep learning models.
Although these solutions run on embedded systems and IoT units, their training is typically carried out on high-performing units (such as Cloud or EdgeComputing systems), with very few papers proposing on-device incremental learning mechanisms~\cite{cai2020tinytl,disabato2020incremental}.

The ability to learn TML models directly on the devices is crucial to improve the accuracy over time by exploiting fresh information coming from the field, and to deal with concept drift, i.e., variations in the statistical behavior of the data generating process, a quite common situation in real-world applications (e.g., due to seasonality or periodicity effects, faults affecting sensors or actuators, changes in the user's behavior, or aging consequences). Failing to adapt TML models to concept drift results in a (possibly dramatic) decrease of the accuracy over time~\cite{ditzler2015learning}.

This paper aims at addressing this challenge by introducing, for the first time in the literature, a Tiny Machine Learning algorithm for Concept Drift (TML-CD) that can learn directly on the IoT unit or embedded system and adapt the knowledge base in response to a concept drift (thus tracking the evolution of the data generating process). To achieve this goal, we introduce three different adaptation mechanisms (i.e., passive, active, and hybrid), each of which has its advantages, issues, accuracy, and behavior. Among these mechanisms, we focus on the hybrid solution thanks to its ability to trade-off adaptation with memory demand. 
The three proposed TML-CD adaptive mechanisms have been tested in two different application scenarios (i.e., image classification and speech command recognition) and ported to three real-world Micro-Controller Units, showing their feasibility and effectiveness. Finally, the code is made available to the scientific community.\footnote{\footnotesize The repository link is https://github.com/simdis/Adaptive-TML.}

The paper is organized as follows. Section~\ref{sct:literature} revises the related literature. Section~\ref{sct:problem_formulation} formalises the addressed problem, whereas Sections~\ref{sct:solution},~\ref{sct:configuration}, and~\ref{sct:adaptation} present the proposed TML-CD solution and its stages. Finally, Section~\ref{sct:experimentalResults} details the experimental results and Section~\ref{sct:conclusions} draws the conclusions.
%
%
\section{Related Literature}
\label{sct:literature}
This section discusses the related literature about machine and deep learning solutions in presence of concept drift as well as TML solutions.
\paragraph*{\textbf{Machine and Deep Learning Solutions in Presence of Concept Drift}}
The literature about machine and deep learning in presence of concept drift refers to adaptive solutions able to deal with concept drift affecting the data generating process. The related literature usually groups them into two main families: passive and active~\cite{ditzler2015learning,gama2014survey,lu2018learning}.

\emph{Passive solutions} adapt the model at each incoming data, disregarding the fact that a concept drift has occurred in the data-generating process (or not). The gradual forgetting classifiers, e.g.~\cite{elwell2011incremental,krawczyk2015one}, which reduce the importance of older samples over time, are examples of passive solutions. The Concept Drift Very Fast Decision Tree (CDVFDT)~\cite{hulten2001mining} introduces a Decision Tree that learns new subtrees on incoming data. However, most passive solutions employ ensemble methods and their adaptation mechanisms consist in adding, removing, or weighting the ensemble base classifiers, e.g., Streaming Ensemble Algorithm~\cite{street2001streaming}, Dynamic Classifier Selection~\cite{almeida2018adapting}, or the adaptive ensemble of Decision Trees proposed in~\cite{pietruczuk2017adjust}. 
Deep learning-based passive solutions examples can be found in~\cite{li2020continual,parisi2019continual,perez2018review}.

On the contrary, \emph{active solutions} aim at detecting concept drift in the data generation process and, only in that case, they adapt their model to the new conditions. Change Detection Tests (CDT) are statistical techniques meant to sequentially process the incoming data inspecting for concept drift.~\cite{ditzler2011hellinger} proposed to use the Hellinger distance between the reference probability distribution and the one estimated on incoming data along with a t-test to detect changes.~\cite{dasu2006information} relies on bootstrapping several windows of data and the Kullback Leibler divergence as a measure of the distance among them. A few works detect changes with density estimation techniques~\cite{bu2016pdf,duda2018parzen}. Other examples of CDT used in active solutions can be found in~\cite{baena2006early,page1954continuous,wang2020multiscale,zambon2018concept}.
In active solutions, the adaptation stage following a concept drift detection is usually carried out in two steps~\cite{alippi2013just}: first, the time instant the concept drift occurred is estimated by ad-hoc mechanisms (e.g., by Change-Point Methods); second, the obsolete knowledge, i.e., that acquired before the concept drift occurred, is discarded. To achieve this goal, the adaptation mechanisms typically rely on a window over the last acquired data, whose size is usually optimized over time to reduce its memory requirements~\cite{aggarwal2006biased,vitter1985random}, or on all the samples seen so far (suitably weighted)~\cite{klinkenberg2004learning}. Finally, deep learning-based active approaches (integrating deep learning solutions with active adaptive solutions) can be found in~\cite{disabato2019learning,yang2019novel}.
\paragraph*{\textbf{Tiny Machine Learning}} 
TML techniques aim at designing machine and deep learning models that take into account the severe technological constraints on memory, computation, and energy characterizing IoT units and embedded systems~\cite{banbury2020benchmarking,disabato2020incremental,sanchez2020tinyml}. 

To achieve this goal, most solutions employ approximation techniques from the deep learning literature. These approximation mechanisms can be grouped into three main families according to the way the approximation is carried out: pruning of processing layers (and part of them)~\cite{he2017channel,lin2018holistic}, quantization of parameters and activations with limited precision or binary parameters~\cite{bulat2021bit,gupta2015deep,rastegari2016xnor}, or solutions integrating both pruning and quantization~\cite{alippi2018moving}.

As regards the target of the approximation mechanisms, most of TML literature focuses on approximated Convolutional Neural Networks~\cite{banbury2021micronets,gopinath2019compiling,kumar2017resource,lin2020mcunet}, with a few works considering recurrent DL architectures~\cite{fedorov2020tinylstms,venzke2020artificial}. In particular,~\cite{fedorov2019sparse} introduces a methodology to explore sparse (and pruned) CNN architectures able to be executed on microcontroller units, whereas~\cite{cai2020tinytl} proposes a Tiny-CNN whose biases can be learned directly on the device. Finally,~\cite{rusci2020leveraging} investigates the impact of quantized networks in TinyML embedded systems.

Although there are very few works proposing on-device learning, e.g.,~\cite{cai2020tinytl,disabato2020incremental}, to the best of our knowledge, no work presents a Tiny-ML solution able to adapt over time to concept drift.
    %
    %
    \section{Problem Formulation}
\label{sct:problem_formulation}
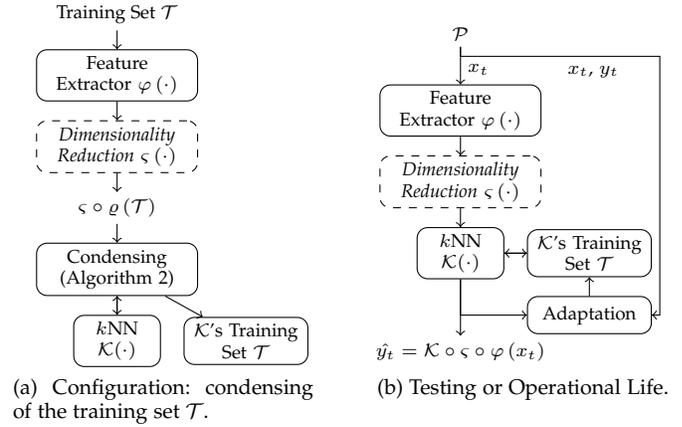
\begin{figure}[!t]
	\centering
    \scriptsize
    \subfloat[][Configuration: condensing of the training set $\mathcal{T}$.]{
        \centering
    \tikzexternalexportnextfalse
    \begin{tikzpicture} [node distance = 1 cm]%
            \centering
            \node (ts) [text centered]
            {\centering Training Set $\mathcal{T}$};
            \node (left_space) [text centered, left=0.25cm of ts]
            {};
            \node (fe) [block, text width=8em, below = 0.25cm of ts] 
            {\centering Feature Extractor $\varphi \left(\cdot\right)$};
            %
            %
            \node (dimred)  [block, dashed, text width=8em, below = 0.25cm of fe]
            {\centering\itshape Dimensionality Reduction $\varsigma \left(\cdot\right)$};
            \node (extracted_ts) [text centered, below=0.25cm of dimred]
            {\centering $\varsigma \circ\varrho\left(\mathcal{T}\right)$};
            \node (condensing) [block, text width=8em, below=0.25 of extracted_ts]
            {\centering Condensing (Algorithm~\ref{algorithm:condensed_nearest_neighbor})};
            \node (knn) [block, text width=4em, below = 0.25cm of condensing] 
            {\centering $k$NN $\mathcal{K}(\cdot)$};
            \node (knn_ts) [block, text width=6em, right=0.3cm of knn]
            {\centering $\mathcal{K}$'s Training Set $\mathcal{T}$};
            %
            %
            %
            %
            \draw[->] (ts) --  (fe);
            \draw [->] (fe) -- (dimred);
            \draw[->] (dimred) -- (extracted_ts);
            \draw[->] (extracted_ts) -- (condensing);
            \draw[->] (condensing) -- (knn_ts);
            \draw[->] (condensing) -- (knn);
            \draw[->] (knn) -- (condensing);
            \label{fig:architectureinitialtraining}
            \end{tikzpicture}
    }
    \hfill
    \subfloat[][Testing or Operational Life.]{
        \centering
        %
        \begin{tikzpicture} [node distance = 1 cm]%
            \centering
            \node (process) [text centered]
            {\centering $\mathcal{P}$};
            \node (right_space) [text centered, right=2.5cm of process]
            {};
            \node (fe) [block, text width=8em, below = 0.5cm of process] 
            {\centering Feature Extractor $\varphi \left(\cdot\right)$};
            %
            %
            \node (dimred)  [block, dashed, text width=8em, below = 0.25cm of fe]
            {\centering\itshape Dimensionality Reduction $\varsigma \left(\cdot\right)$};
            \node (knn) [block, text width=4em, below = 0.25cm of dimred] 
            {\centering $k$NN $\mathcal{K}(\cdot)$};
            \node (knn_ts) [block, text width=6em, right=0.3cm of knn]
            {\centering $\mathcal{K}$'s Training Set $\mathcal{T}$};
            \node (output)  [text centered, below=0.75cm of knn] {\scriptsize$\hat{y_t} = \mathcal{K}\circ \varsigma \circ \varphi \left( x_t\right)$};
            \node (adaptation) [block, text width=6em, below=0.25cm of knn_ts]
            {\centering Adaptation};
            %
            %
            %
            %
            %
            \draw[->] (process) --
            node[text width=1cm,xshift=0.6cm, yshift=-0.05cm] {\scriptsize\centering $x_t$}
            (fe);
            \draw [->] (fe) -- (dimred);
            \draw [->] (dimred) -- (knn);
            \draw[->] (knn) -- (output);
            \draw[->] (knn) |- (adaptation);
            \draw[<-] (knn) -- (knn_ts);
            \draw[->] (knn) -- (knn_ts);
            \draw[->] (process) -- ++(0, -0.3cm) -- node[text width=1cm,xshift=0.6cm, yshift=-0.2cm] {\scriptsize\centering $x_t$, $y_t$}
            ++(2.65cm, 0) |- (adaptation);
            \draw[->] (adaptation) -- (knn_ts);
            %
            %
            \label{fig:architecturetest}
        \end{tikzpicture}
    }
	%
    %
	\caption{The proposed architecture of the proposed solution for Tiny Machine Learning for Concept Drift (TML-CD)~$\mathcal{D}$ on embedded systems and IoT units.}
	%
    %
	\label{fig:architecture}
\end{figure}
Let $\mathcal{P}$ be a data generating process that, at each time instant $t$, provides a pair $\left(x_t, y_t\right)$ sampled from an unknown probability distribution $p_t(x,y)$, where $x$ is the input of the proposed solution $\mathcal{D}$ (e.g., an image or an audio clip) and $y\in \Lambda$ its classification label.\footnote{Let $\lambda$  be the cardinality of $\Lambda$, i.e., the number of classes in the considered classification problem.}

Moreover, following a \emph{test-then-train} approach~\cite{ditzler2015learning}, the proposed solution $\mathcal{D}$ receives the supervised information (the true label $y_t$) only after it provides the classification output $\hat{y}_t = \mathcal{D}\left(x_t\right)$ on input $x_t$, at each time instant $t$.\footnote{Without any loss of generality, the supervised information might not be provided at every time instant. In those cases, the proposed solution only provides its classification output.} 

In a concept drift scenario, the process $\mathcal{P}$ might evolve over time, hence inducing a shift in the distribution $p_t(x,y)$ at an unknown time instant $t^{*}$. It is worth noting that the change in $p_t(x,y)$ might affect the input $x$ (e.g., by the introduction of noise), the set $\Lambda$ (e.g., class change), or both. 

The goal of the proposed TML-CD solution is to react and adapt $\mathcal{D}$  to changes in $p_t(x,y)$ so as to guarantee the highest accuracy over time.
\section{The Proposed Tiny Machine Learning for Concept Drift }
\label{sct:solution}
\begin{algorithm}[t]
    \DontPrintSemicolon
    \small
    \KwInput{Training Set $\mathcal{T}$, Feature Extractor $\varsigma \circ\varrho$.}
    \KwParameter{Number of neighbors $k$.}
    Preprocess $\mathcal{T}$.\tcc*{Initialization.}
    \label{alg_line:start_initialization}
    \label{alg_line:preprocessing}
    Initialize the $k$-NN classifier $\mathcal{K}$ with $\varsigma \circ\varrho\left(\mathcal{T}\right)$.\;
    Define $\mathcal{D} = \mathcal{K} \circ \varsigma \circ\varrho$.\;
    \label{alg_line:end_initialization}
    \tcc*{Loop over samples arriving at time $t$.}
    \ForEach{$\left(x_t, y_t\right) \sim \mathcal{P}, t=1,2,\ldots$ }{
        \label{alg_line:start_online_step}
        Predict $\hat{y_t} \leftarrow \mathcal{D}(x_t)$.\;
        \label{alg_line:prediction}
        \textit{Adaptation of $\mathcal{D}$}.\;
        \label{alg_line:end_online_step}
        \label{alg_line:incremental_update}
    }
    \caption{\small A sketch of the proposed solution.}
    \label{algorithm:sketch}
\end{algorithm}
Figure~\ref{fig:architecture} shows the general architecture of the proposed solution for Tiny Machine Learning for Concept Drift (TML-CD) $\mathcal{D}$, which comprises the following five different modules:
\begin{itemize} 
    \item {\textbf{Feature Extractor} \boldmath$\varrho$}.
    The \emph{Feature Extractor} extracts features from the input $x_t$. As in~\cite{alippi2018moving,disabato2020incremental}, the Feature Extractor is a pre-trained DL model approximated by means of Task-Dropping (e.g., pruning of layers), Precision Scaling (e.g., weights precision reduction), or both, to satisfy the constraints on computation, memory, and energy characterizing the embedded systems and IoT units running $\mathcal{D}$.
    \item {\textbf{Dimensionality Reduction Operator} \boldmath$\varsigma$}
    The \emph{Dimensionality Reduction Operator} $\varsigma$ (that can be optionally activated) reduces the dimensionality of features extracted by $\varrho$. In this paper, among the approaches presented in~\cite{disabato2020incremental}, we focused on the \emph{Filter-Selection without supervised information}. This technique selects the $f$ out of $F$ filters of the last $\varrho$ convolutional layer (and its subsequent batch-normalization channels, if any) providing the highest mean activation on publicly available benchmarks or datasets.
    
    It is crucial to point out that the adaptation step of $\mathcal{D}$ does not affect the feature extractor $\varrho$ nor the dimensionality reduction operator $\varsigma$, which are therefore fixed over time. Moreover, since the choice of the $f$ filters to keep does not rely on the specific data-generation process $\mathcal{P}$, the block $\varsigma \circ \varrho$ can be defined at design time and prior to the porting of $\mathcal{D}$ on the IoT units. This is the reason why $\varsigma \circ \varrho$ is an input to our algorithm, and $\varsigma$ takes part in the choice of the approximated DL-feature extractor.
    \item {\boldmath$k$\textbf{NN Classifier} and \textbf{Training Set} $\mathcal{T}$}.
    The $k$NN~\cite{altman1992introduction} \emph{classifier} $\mathcal{K}\left(\cdot\right)$, whose input is either the output of the dimensionality-reduction operator  $\varsigma \circ \varrho$ or that of the feature extractor $\varrho$ (when no dimensionality reduction is considered), provides the classification $\hat{y_t}$ of the input $x_t$, while $\mathcal{T}$ is its training set. From the algorithmic point of view, the $k$NN is a statistical classifier based on majority voting, i.e., the predicted class corresponds to the majority class of the $k$ nearest neighbors of the input sample within $\mathcal{K}$'s training set $\mathcal{T}$. Interestingly, it does not require a training phase, but only the initialization of its training set $\mathcal{T}$. Unless otherwise specified, the parameter $k$, i.e., the number of neighbors, is set to the ceiling of the square root of the available samples, as suggested in~\cite{alippi2013just}.
    \item \textbf{Adaptation Module}.
    The adaptation module receives as input the sample $x_t$ and its $k$NN $\mathcal{K}$ prediction $\hat{y_t}$ and, when the supervised information $y_t$ is available, it updates the TML-CD solution $\mathcal{D}$ so as to make it adaptive over time to concept drift. Among the four presented modules, the adaptation involves only the $\mathcal{K}$'s training set $\mathcal{T}$~\cite{losing2016knn,roseberry2018multi}. The $k$NN classifier adaption indeed requires to simply add the new supervised information $\left(x_t, y_t\right)$  to its training set $\mathcal{T}$.
\end{itemize}
Algorithm~\ref{algorithm:sketch}, instead, details how the proposed TML-CD $\mathcal{D}$ works.
More in detail, the TML-CD $\mathcal{D}$, which receives in input a feature extractor along with a dimensionality reduction operator ($\varsigma \circ \varrho$) and an initial training set $\mathcal{T}$, comprises two different stages: configuration and testing. 

The \emph{configuration} stage, detailed in  lines~\ref{alg_line:start_initialization}--\ref{alg_line:end_initialization} and shown in Figure~\ref{fig:architectureinitialtraining}, encompasses an initial \emph{preprocessing step} where the training set $\mathcal{T}$ is preprocessed to reduce the memory occupation (line~\ref{alg_line:preprocessing}) by means of a condensing mechanism (Algorithm~\ref{algorithm:condensed_nearest_neighbor}). Once the preprocessing step has been carried out, the knowledge base of the $k$NN classifier is initialized on the features extracted from the preprocessed training set $\mathcal{T}$, i.e., the training set of $\mathcal{K}$ is $\varsigma \circ\varrho\left(\mathcal{T}\right)$. Section~\ref{sct:configuration} will detail the configuration stage.

After the completion of the configuration stage, the TML-CD solution $\mathcal{D} = \mathcal{K} \circ \varsigma \circ\varrho$ enters the \emph{testing} stage where it is able to operate on the novel incoming samples provided by the data-generating process $\mathcal{P}$ (lines~\ref{alg_line:start_online_step}--\ref{alg_line:end_online_step}). At each time instant $t=1,2,\ldots$, the proposed solution $\mathcal{D}$ receives in input $x_t$ and provides the output $\hat{y_t} = \mathcal{D}(x_t)$ (line~\ref{alg_line:prediction}). Then, when the supervised information $y_t$ about $x_t$ is made available as per the "test-and-train" approach, it activates the  adaptation step (line~\ref{alg_line:incremental_update}). Section~\ref{sct:adaptation}  will detail the testing stage by describing the proposed three adaptive mechanisms for TML-CD.
\section{The configuration stage: condensing $\mathcal{T}$}
\label{sct:configuration}
\begin{algorithm}[t]
    \DontPrintSemicolon
    \small
    \KwInput{Training Set $\mathcal{T}$.}
    \KwOutput{Condensed Representation $\bar{\mathcal{T}} \subseteq \mathcal{T}$.}
    \tcc*{$H$ contains one sample, $D$ all the others.}
    Initialize $H\leftarrow\left\{t \in \mathcal{T}\right\}$ and $D\leftarrow\mathcal{T} \setminus H$.\;
    Initialize the kNN $\mathcal{K}$ with $H$.\;
    \Do{$H$ and $D$ are modified in the \textit{foreach} loop.}{
            \ForEach{$t\leftarrow\left(x, y\right) \in D$}{
                    Predict $\hat{y} \leftarrow \mathcal{K}(t)$.\;
                    \If(\tcc*[f]{Condensing Update:}){$\hat{y} \not= y$}{
                          \label{alg_line:condensing_update}
                           $ D \leftarrow D \setminus \{t\} $. \tcc*{Move $t$ from $D$ to $H$.}
                            $ H \leftarrow H \cup \{t\}$.  \;
                            (Re--)train the kNN $\mathcal{K}$ with $H$.\;
                            \label{alg_line:end_condensing_update}
                    }    
            }
    }
    \Return{$\bar{\mathcal{T}} \leftarrow H$}
    \caption{\small The Condensed Nearest Neighbor~\cite{hart1968condensed}.}
    \label{algorithm:condensed_nearest_neighbor}
\end{algorithm}
The $k$NN classifier has the great advantage of not requiring a proper training phase. However, this advantage comes, in principle, at the expense of the following two drawbacks. 
First, a $k$NN-based classifier requires to store all the data of the training set. Second, the larger the amount of the training data, the higher the time to provide a classification in output. These drawbacks are more severe as the samples within $\mathcal{T}$ increase.

The related literature addresses these two issues from three different perspectives. 

First, condensing techniques~\cite{hart1968condensed,tomek1976two} aim at identifying the smallest subset of training data that can correctly classify all the training samples. Second, editing techniques~\cite{laurikkala2001improving,tomek1976an,wilson1972asymptotic} instead reduce the number of stored samples by removing the noisy ones, i.e., those not agreeing with their neighborhoods. Third,~\cite{smith2014instance} proposed to train a supervised parametric classifier on available data and to remove all the samples having a classification probability below a hard threshold. 

In this work, we focus on the first approach and, in particular, on the Condensed Nearest Neighbour algorithm. In more detail, $\mathcal{D}$ applies this algorithm during the preprocessing step (Algorithm~\ref{algorithm:sketch}--Line~\ref{alg_line:preprocessing}) in order to optimize both the memory and computational requirements of the classifier $\mathcal{K}$. More specifically, given a $N$-dimensional training set $\mathcal{T} = \left\{\left(x_t, y_t\right), t=1,\ldots,N\right\}$, the preprocessing step computes the condensed representation of $T$, i.e., the minimum subset $\bar{\mathcal{T}} \subseteq \mathcal{T}$ for which $\mathcal{K}$ is able to correctly classify all the samples in $\mathcal{T}$. Algorithm~\ref{algorithm:condensed_nearest_neighbor} shows the pseudo-code of the condensing algorithm proposed by Hart~\cite{hart1968condensed} that is employed in the preprocessing step.
Section~\ref{subsct:condensing_impact} experimentally evaluates the impact on accuracy and memory demand of this condensing stage, highlighting that the significant savings in terms of memory come at the expense of a negligible drop in accuracy (in stationary conditions).
\section{The testing stage: adapting $\mathcal{T}$} 
\label{sct:adaptation}
The adaptation module, which is the core of the proposed $\mathcal{D}$, has been declined from three different perspectives, differing in the type of adaptation mechanism therein employed: 
\begin{itemize}
    \item \textbf{Passive Update} (Section~\ref{sct:condensing_in_time}). The adaptation module relies on a fully passive approach where the adaptation is carried out at each new incoming supervised samples, without requiring an explicit detection of a change in the data-generating process $\mathcal{P}$;
    \item \textbf{Active Update} (Section~\ref{sct:active_approach}). This adaptation module relies on a CDT to detects changes in $\mathcal{P}$. Once a change is detected, the algorithm adapts $\mathcal{K}$ accordingly;
    \item \textbf{Hybrid Update} (Section~\ref{sct:hybrid_approach}). The hybrid adaptation module integrates the passive approach with a CDT to speed up the adaptation stage exactly when needed.
\end{itemize} 
\subsection{Passive Update: the Condensing-in-Time approach}
\label{sct:condensing_in_time}
\begin{algorithm}[t]
    \DontPrintSemicolon
    \small
    \KwInput{Training Set $\mathcal{T}$, Feature Extractor $\varsigma \circ\varrho$.}
    \KwParameter{Maximum number of training samples $p$.}
    Compute $\mathcal{T} \leftarrow \bar{\mathcal{T}}$ with Algorithm~\ref{algorithm:condensed_nearest_neighbor}.\tcc*{Condense $\mathcal{T}$.}
    Initialize the $k$-NN classifier $\mathcal{K}$ with $\varsigma \circ\varrho\left(\mathcal{T}\right)$.\;
    Define $\mathcal{D} = \mathcal{K} \circ \varsigma \circ\varrho$.\;
    \tcc*{Loop over samples arriving at time $t$.}
    \ForEach{$\left(x_t, y_t\right) \sim \mathcal{P}, t=1,2,\ldots$ }{
                  Predict $\hat{y_t} \leftarrow \mathcal{D}(x_t)$\;
                        
                        \If(\tcc*[f]{Passive Update.}){$\hat{y_t} \not= y_t$}{
                            \label{alg_line:start_cit_passive_update}
                            $\mathcal{T} \leftarrow \mathcal{T} \cup \left(x_t, y_t\right)$\;
                            \If(\tcc*[f]{Window Size Check.}){$\left|\mathcal{T}\right| > p$}{
                                    \label{alg_line:start_cit_window_size_check}
                                    $\left(x_{\tilde{t}}, y_{\tilde{t}}\right) \leftarrow \argmin_{\bar{t}} \left\{\left(x_{\bar{t}}, y_{\bar{t}}\right) \in \mathcal{T}\right\}$\;
                                    $\mathcal{T} \leftarrow \mathcal{T} \setminus \left\{\left(x_{\tilde{t}}, y_{\tilde{t}}\right)\right\}$\;
                                    \label{alg_line:end_cit_window_size_check}
                            }
                            Update $\mathcal{D}$ with $\mathcal{T}$\;
                            \label{alg_line:end_cit_passive_update}
                        }
     }
    \caption{\small The Condensing-in-Time (Passive).}
    \label{algorithm:condensing_in_time}
\end{algorithm}
The passive approach, called \emph{Condensing-in-Time} (CIT) algorithm, updates the training set $\mathcal{T}$ every time a new supervised sample is available. Algorithm~\ref{algorithm:condensing_in_time} presents the CIT algorithm. 

It receives in input the feature extractor along with a dimensionality reduction operator $\varsigma \circ \varrho$ and a training set $\mathcal{T}$, whose condensed representation $\bar{\mathcal{T}}\subseteq \mathcal{T}$ (see Algorithm~\ref{algorithm:condensed_nearest_neighbor}) is used to initialize the training set $\mathcal{T}$ of the $k$NN. Once initialized, the CIT--$\mathcal{D}$, i.e.,  the $\mathcal{D}$ solution implementing the CIT adaption module, is ready to classify novel incoming samples. The CIT passively updates the $\mathcal{K}$'s knowledge set $\mathcal{T}$ at every time instant $t$ for which the supervised information, i.e., the true label $y_t$, is available. More in details, CIT--$\mathcal{D}$ adds the sample $x_t$ and its true label $y_t$ (at time instant $t$) to the $k$NN $\mathcal{K}$ knowledge set $\mathcal{T}$ if and only if $x_t$ is misclassified, i.e., $\hat{y_t} \not= y_t$  (Algorithm~\ref{algorithm:condensing_in_time}, Lines~\ref{alg_line:start_cit_passive_update}--\ref{alg_line:end_cit_passive_update}). This idea is inspired by the condensing algorithm update, shown at Lines~\ref{alg_line:condensing_update}--\ref{alg_line:end_condensing_update} in Algorithm~\ref{algorithm:condensed_nearest_neighbor}, but  it is here tailored to the time evolution of the data-generating process. 

It is worth noting that the CIT algorithm can only add a new supervised sample to the knowledge set $\mathcal{T}$ of the $k$NN $\mathcal{K}$, hence potentially introducing critical issues in the memory and computational demand of the $k$NN when the number of samples in $\mathcal{T}$ increases.
To keep under control the cardinality of $\mathcal{T}$, the CIT algorithm employs two different solutions. 
The former introduces a maximum number of samples $p$ that can be stored, i.e., $\left|\mathcal{T}\right| \le p$, being $\left|\cdot\right|$ cardinality operator. Hence, every time the adaptation stage introduces a sample in $\mathcal{T}$ overcoming this limit, the oldest sample is removed (Algorithm~\ref{algorithm:condensing_in_time}, Lines~\ref{alg_line:start_cit_window_size_check}--\ref{alg_line:end_cit_window_size_check}). As a consequence, the solution $\mathcal{D}$ based on the CIT classifier operates on the last $p$ supervised samples introduced in $\mathcal{T}$. Besides, this mechanism allows also to remove old samples in $\mathcal{T}$ by introducing only misclassified samples, i.e., those bringing more information to the classifier. 

The latter introduces a probability for a misclassified sample to be added to the training set $\mathcal{T}$. Ideally, such probability should be close to zero in stationary conditions and close to one immediately after a change. The definition of this probabilistic memory management mechanism is left as future work.
\subsection{Active Update: Active Tiny $k$NN}
\label{sct:active_approach}
\begin{algorithm}[t]
    \DontPrintSemicolon
    \small
    \KwInput{Feature Extractor $\varsigma \circ\varrho$, CDT $\vartheta$, Training Set $\mathcal{T}$.}
    \KwParameter{History Window Size $\varpi$, CDT threshold $h$.}
    Compute $\mathcal{T} \leftarrow \bar{\mathcal{T}}$ with Algorithm~\ref{algorithm:condensed_nearest_neighbor}.\tcc*{Condense $\mathcal{T}$.}
    Initialize the $k$-NN classifier $\mathcal{K}$ with $\varsigma \circ\varrho\left(\mathcal{T}\right)$.\;
    Define $\mathcal{D} = \mathcal{K} \circ \varsigma \circ\varrho$.\;
    Initialize $W \leftarrow \emptyset$.\tcc*{History Window.}
    \tcc*{Loop over samples arriving at time $t$.}
    \ForEach{$\left(x_t, y_t\right) \sim \mathcal{P}, t=1,2,\ldots$ }{
        Predict $\hat{y_t} \leftarrow \mathcal{D}(x_t)$.\;
        $W \leftarrow W \cup \left\{\left(x_t, y_t\right)\right\}$.\tcc*{Update History Window.}
         \label{alg_line:start_update_history_window}
        \label{alg_line:start_active_step}
        \If{$\left|W\right| \ge \varpi$}{
                $W \leftarrow W \setminus \left\{\left(x_{t-\varpi}, y_{t-\varpi}\right)\right\}$.\;
        }
    \label{alg_line:end_update_history_window}
        Compute CDT metric $s_t$.\tcc*{Active Step.}
        \label{alg_line:start_cdt_step}
        Apply CDT $g_t\leftarrow \vartheta\left(s_1,\ldots,s_t\right)$.\;
        \label{alg_line:end_cdt_step}
        \If(\tcc*[f]{Change Detection Check.}){$g_t \ge h$}{
               \label{alg_line:cdt_detection_test}
                Estimate Real Change Time $t_r$.\;
                \label{alg_line:start_adaptation}
                $\mathcal{T}\leftarrow \left\{ \left(x_{\bar{t}}, y_{\bar{t}}\right) \in W : \bar{t} \ge t_r\right\}$.\tcc*{Novel Samples.}
                \textit{[Optional] Condense $\mathcal{T}$.}\;
                Update $\mathcal{D}$ with $\mathcal{T}$.\;
                \label{alg_line:end_adaptation}
                \label{alg_line:end_active_step}    
        }
    }
    \caption{\small The Active Tiny $k$NN.}
    \label{algorithm:active}
\end{algorithm}
The Active Tiny $k$NN, whose pseudocode is shown in Algorithm~\ref{algorithm:active}, relies on a Change Detection Test (CDT) $\vartheta$ to detect changes in the data generation process $\mathcal{P}$. The core of this algorithm is the ability to adapt the classifier $\mathcal{K}$'s training set $\mathcal{T}$ only after the detection of a concept drift. In addition, the Active Tiny $k$NN allocates space for a history window $W$ of size $\varpi$, being $\varpi$ a parameter of the algorithm (described in the sequel).

In more detail, for each sample $\left(x_t, y_t\right)$ provided by  $\mathcal{P}$ at time instant $t$, the Active Tiny $k$NN predicts the label $\hat{y_t} = \mathcal{D} (x_t)$ and, when the supervised information $y_t$ is available, the active update is activated (Algorithm~\ref{algorithm:active}, Lines~\ref{alg_line:start_active_step}--\ref{alg_line:end_active_step}). 

At first, it adds the pair $\left(x_t, y_t\right)$ to the history window $W$ and discards the oldest pair if the window already contains $\varpi$ pairs (Algorithm~\ref{algorithm:active}, Lines~\ref{alg_line:start_update_history_window}--\ref{alg_line:end_update_history_window}). After that, the Active Tiny $k$NN computes the figure of merit $s_t$ (at time $t$) and applies the CDT decision function $\vartheta$ to inspect for changes in $\mathcal{P}$ (Algorithm~\ref{algorithm:active}, Lines~\ref{alg_line:start_cdt_step}--\ref{alg_line:end_cdt_step}), i.e.,  $g_t = \vartheta\left(s_1,\ldots,s_t\right)$. In the most general situation, the computation of $g_t$ at time $t$ takes into account all the figures of merits computed from $t=1$. A change is detected in the data-generation process $\mathcal{P}$ when $g_t$ overcomes the detection threshold $h$, being $h$ a parameter of the algorithm (Algorithm~\ref{algorithm:active}, Line~\ref{alg_line:cdt_detection_test}).

Once a change is detected, the adaptation stage starts (Algorithm~\ref{algorithm:active}, Lines~\ref{alg_line:start_adaptation}--\ref{alg_line:end_adaptation}). In the first place, it estimates the time $t_r$ the change occurred at (e.g., with a Change Point Method). After that, it discards from the history window $W$ all the samples older than the estimated change time $t_r$. The updated history window $W$ (optionally condensed through Algorithm~\ref{algorithm:condensed_nearest_neighbor}) becomes the new $\mathcal{K}$'s training set $\mathcal{T}$.

It is noteworthy to point out that the memory footprint of the Active Tiny $k$NN is bounded over time since it requires to store the training set $\mathcal{T}$ and history window $W$ of at most $\varpi$ samples (the CDT memory footprint can be neglected). Moreover, since the adaptation stage modifies the knowledge set $\mathcal{T}$ only through copies of the (at most whole) history window $W$, the total memory footprint cannot overcome twice the memory of the history window $W$, i.e., that of $2\varpi$ samples.

Although the solution accepts as input any CDT $\vartheta$, in the context of this paper, $\vartheta$ is the well-known and theoretically grounded CUSUM algorithm~\cite{page1954continuous} in its generalized version~\cite{lorden1971procedures}, monitoring the accuracy of the Active Tiny $k$NN over time. As a consequence, any change in the data-generation process $\mathcal{P}$ is assumed to reflect on the $\mathcal{K}$ classification accuracy.

The generalized CUSUM CDT is designed as follows. Let $\upsilon_0$ be the stationary classification accuracy (estimated on the first $\xi$ supervised samples in the testing stage, being $\xi$ parameter of the Active Tiny $k$NN algorithm). A Bernoulli distribution with parameter $\upsilon_0$ and, in turn, a Binomial distribution with parameters $\upsilon_0$ and $n$ (with $n$ size of the batches on which the accuracy is computed in the following) model our scenario in stationary conditions.
The figure of merit of the CUSUM CDT is the likelihood ratio of the probability distributions modeling the scenario after and before the change, i.e.:
\begin{equation}
    s_t = \ln \frac{p_{\upsilon_1}(\zeta_t)}{p_{\upsilon_0}(\zeta_t)},
    \label{eq:log_likelihood_ratio}
\end{equation}
where $\upsilon_1$ represents the classification accuracy after the change and $\zeta_t$ the realization of the Binomial distribution at time $t$, i.e., the accuracy on the $n$ supervised samples arrived before time $t$.\footnote{Although the Active Tiny $k$NN algorithm is general enough to deal with any CDT, in the described CUSUM case with Binomial distribution of size $n$, the CDT figure of merit is not computed for every supervised samples, but every $n$. As a consequence, all the $n-1$ $s_t$ values before a window of size $n$ is full are set to zero.}

Since the value of $\upsilon_1$ is a priori unknown, the generalized version of the CUSUM algorithm employs a set $\Upsilon_1$ containing a grid of possible accuracies $\upsilon_1$ after change equally spaced from $0$ to $1$, except for the neighborhood of $\upsilon_0$.\footnote{The cardinality of $\Upsilon_1$, i.e., the number of tested values $\upsilon_1$, is a parameter of the Active Tiny $k$NN.} The resulting decision function $\vartheta$ is:
\begin{equation}
    g_t = \vartheta\left(s_1, \ldots, s_t\right) = \max_{1 \le j \le t}\ \sup_{\upsilon_1 \in \Upsilon_1} S_j^t (\upsilon_1),
    \label{eq:cusum_decision_function}
\end{equation}
where
\begin{equation}
    S_j^t (\upsilon_1) = \sum_{i=j}^{t} s_i,
    \label{eq:sum_log_likelihood_ratios}
\end{equation}
represents the sum of the log-likelihood ratio $s_t$s from time $j$ to time $t$.

Assuming the parameter $n$ large enough, the considered Binomial distribution can be approximated as a Normal one with mean $n\upsilon_0$ and variance $n\upsilon_0(1-\upsilon_0)$. Hence, the log-likelihood ratio $s_t$ in Equation~\ref{eq:log_likelihood_ratio} then becomes:
\begin{multline}
    s_t = \frac{\upsilon_1\bar{\upsilon_1}-\upsilon_0\bar{\upsilon_0}}{2n\upsilon_0\bar{\upsilon_0}\upsilon_1\bar{\upsilon_1}} \zeta_t^2 + \frac{\bar{\upsilon_0} - \bar{\upsilon_1}}{\bar{\upsilon_0}\bar{\upsilon_1}} \zeta_t+ \\ +\frac{n(\upsilon_0\bar{\upsilon_1}-\upsilon_1\bar{\upsilon_0})}{2\bar{\upsilon_0}\bar{\upsilon_1}} + \ln\sqrt{\frac{\upsilon_0\bar{\upsilon_0}}{\upsilon_1\bar{\upsilon_1}}},
    \label{eq:binomial_log_likelihood_ratio}
\end{multline}
where $\bar{\upsilon_0} = 1 - \upsilon_0$ and $\bar{\upsilon_1} = 1 - \upsilon_1$.

As a final remark, the CUSUM CDT is also endowed with the ability to estimate the change time $t_r$ (and, if desired, of the parameter $\upsilon_1$ after the change). The estimated change time $t_r$ is indeed the index $\tilde{j}$ maximizing the decision function $\vartheta$ in Eq.~\eqref{eq:cusum_decision_function}, i.e.:
\begin{equation}
    \left(\tilde{j}, \tilde{\upsilon_1}\right) = \argmax_{1 \le j \le t}\ \sup_{\upsilon_1 \in \Upsilon_1} S_j^t (\upsilon_1).
    \label{eq:change_point_method}
\end{equation}
\subsection{Hybrid Tiny $k$NN: Integrating Condensing-in-Time and Active Tiny $k$NN}
\label{sct:hybrid_approach}
\begin{algorithm}[t]
    \DontPrintSemicolon
    \small
    \KwInput{Feature Extractor $\varsigma \circ\varrho$, CDT $\vartheta$, Training Set $\mathcal{T}$.}
    \KwParameter{Maximum $\mathcal{T}$ Size $\varpi$, CDT threshold $h$.}
    Compute $\mathcal{T} \leftarrow \bar{\mathcal{T}}$ with Algorithm~\ref{algorithm:condensed_nearest_neighbor}.\tcc*{Condense $\mathcal{T}$.}
    Initialize the $k$-NN classifier $\mathcal{K}$ with $\varsigma \circ\varrho\left(\mathcal{T}\right)$.\;
    Define $\mathcal{D} = \mathcal{K} \circ \varsigma \circ\varrho$.\;
    \tcc*{Loop over samples arriving at time $t$.}
    \ForEach{$\left(x_t, y_t\right) \sim \mathcal{P}, t=1,2,\ldots$ }{
        Predict $\hat{y_t} \leftarrow \mathcal{D}(x_t)$.\;
        \If(\tcc*[f]{Passive Update.}){$\hat{y_t} \not= y_t$}{
            \label{alg_line:start_hybrid_passive_update}
            $\mathcal{T} \leftarrow \mathcal{T} \cup \left(x_t, y_t\right)$\;
            \If{$\left|\mathcal{T}\right| \ge \varpi$}{
                $t_{min} \leftarrow \min_{\bar{t}} \left\{ \left(x_{\bar{t}}, y_{\bar{t}}\right) \in \mathcal{T}\right\}$\;
                $\mathcal{T} \leftarrow \mathcal{T} \setminus \left\{\left(x_{t_{min}}, y_{t_{min}}\right)\right\}$.\;
            }
            Update $\mathcal{D}$ with $\mathcal{T}$.\;
        }
        \label{alg_line:end_hybrid_passive_update}
        Compute CDT metric $s_t$.\tcc*{Active Step.}
        \label{alg_line:start_hybrd_active_step}
        \label{alg_line:start_hybrid_cdt_step}
        Apply CDT $g_t\leftarrow \vartheta\left(s_1,\ldots,s_t\right)$.\;
        \label{alg_line:end_hybrid_cdt_step}
        \If(\tcc*[f]{Change Detection Check.}){$g_t \ge h$}{
            \label{alg_line:cdt_hybrid_detection_test}
            Estimate Real Change Time $t_r$.\;
            \label{alg_line:start_hybrid_adaptation}
            $\mathcal{T}\leftarrow \left\{ \left(x_{\bar{t}}, y_{\bar{t}}\right) \in \mathcal{T} : \bar{t} \ge t_r\right\}$.\tcc*{Novel Samples.}
            \textit{[Optional] Condense $\mathcal{T}$.}\;
            Update $\mathcal{D}$ with $\mathcal{T}$.\;
            \label{alg_line:end_hybrid_adaptation}
            \label{alg_line:end_hybrid_active_step}    
        }
    }
    \caption{\small The Hybrid Tiny $k$NN.}
    \label{algorithm:hybrid}
\end{algorithm}
The core of the proposed hybrid update is to integrate the "condensing-in-time" ability of the passive update with the capability to quickly adapt to changes by discarding obsolete knowledge of the active one. 

In more detail, the (CIT) passive update continuously adapts $\mathcal{T}$ when supervised information is available, regardless a concept drift occurred (or not). This ability comes at the expense of two weak points. First, there is (in principle) no bound on the memory occupation, although two solutions have been suggested to mitigate the problem. Second, when a change occurs, the passive update does not discard the obsolete knowledge present in $\mathcal{T}$, i.e., samples generated by $\mathcal{P}$ before the concept drift occurred.

On the contrary, the active adaptation provides a bound on the memory occupation (i.e., twice the history window size $W$) and, in turn, on the required computation. However, similarly to the other active approaches present in the literature~\cite{baena2006early,disabato2019learning,wang2020multiscale}, the effectiveness of the active adaptation phase is strictly related to the ability to promptly detect the concept drift in $\mathcal{P}$. 

The proposed hybrid update aspires at compensating the weak points of passive and active updates by integrating the "condensing-in-time" solution described in Algorithm~\ref{algorithm:condensed_nearest_neighbor} with the CUSUM-based CDT detailed in Section~\ref{sct:active_approach}. The resulting algorithm, namely the Hybrid Tiny $k$NN, is shown in Algorithm~\ref{algorithm:hybrid}. Here, the inputs and the initialization are the same as Active Tiny $k$NN. The only difference resides in the fact that the Hybrid Tiny $k$NN does not allocate a history window, but it relies on the training set $\mathcal{T}$ (whose size is bounded by $\varpi$) as history window.

Similarly to the algorithms it derives from, the Hybrid Tiny $k$NN predicts the label $\hat{y_t} = \mathcal{D} (x_t)$ and, when the supervised information $y_t$ is made available, it carries out both a passive (Algorithm~\ref{algorithm:hybrid}, Lines~\ref{alg_line:start_hybrid_passive_update}--\ref{alg_line:end_hybrid_passive_update}) and an active update (Algorithm~\ref{algorithm:hybrid}, Lines~\ref{alg_line:start_hybrd_active_step}--\ref{alg_line:end_hybrid_active_step}), for each sample $\left(x_t, y_t\right)$ generated by  $\mathcal{P}$ at time instant $t$. Although the passive update is equal to that of the "condensing-in-time" algorithm, the active one requires to take into account the effects of the passive updates, which are supposed to increase the classification capability of the algorithm over time  (until the accuracy of Hybrid Tiny $k$NN reaches its maximum value). Consequently, the CUSUM CDT is slightly modified in its set $\Upsilon_1$, which contains only values that are smaller than $\upsilon_0$, i.e., the accuracy estimated on an initial window of data. In this way, the hybrid update does not detect as concept drift the increases in the accuracy brought by the passive update (hence focusing on changes inducing a drop in the accuracy). Moreover, the adaptation phase triggered by the CDT involves directly the knowledge set $\mathcal{T}$ of  the classifier $\mathcal{K}$, where samples older than the estimated time of change $t_r$ are discarded (Algorithm~\ref{algorithm:hybrid}, Lines~\ref{alg_line:start_hybrid_adaptation}--\ref{alg_line:end_hybrid_adaptation}).

Summing up, the hybrid update continuously adapts $\mathcal{T}$ over time thanks to the passive adaptation, hence avoiding the risk of non-detecting changes due to false-negative detections of the CDT. At the same time, the active adaptation present in the hybrid update can quickly discard obsolete knowledge when a change is detected and set a bound on the memory footprint of $\mathcal{T}$.
    %
    %
    \section{Experimental Results}
\label{sct:experimentalResults}
\begin{table*}[t]
    \scriptsize
    \setlength{\tabcolsep}{3pt}
    \centering
    \caption{The impact of condensing techniques (C) in both the application scenarios and in stationary conditions, when no update is done. A SVM and Neural Network (with one fully-connected layer) classifiers have been added as baselines. The results represent the mean $\pm$ standard deviation accuracy ($\alpha_\lambda$) and memory ($m_\lambda$) over 20 experiments with $\lambda=\{2,3,5\}$ classes. The memory is expressed in terms of number of training samples within $\mathcal{T}$, with $\left|\mathcal{T}\right|=100\cdot\lambda$.}
    \begin{tabular}{@{}m{0pt}@{}C{1.75cm}cC{1.25cm}C{1cm}C{1.25cm}C{1cm}C{1.25cm}C{1cm}cC{1.25cm}C{1cm}C{1.25cm}C{1cm}C{1.25cm}C{1cm}@{}m{0pt}@{}}
        &&&\multicolumn{6}{c}{Speech Command Identication}&&\multicolumn{6}{c}{Image Classification}&\\ \cmidrule{4-9} \cmidrule{11-17}
        
        
        &Algorithm&&$\alpha_2$&$m_2$&$\alpha_3$&$m_3$&$\alpha_5$&$m_5$&&$\alpha_2$&$m_2$&$\alpha_3$&$m_3$&$\alpha_5$&$m_5$& \\ \cmidrule{1-2} \cmidrule{4-9} \cmidrule{11-17}
        &SVM&&\textbf{0.93$\pm$0.05}&138$\pm$17&\textbf{0.88$\pm$0.04}&244$\pm$16&\textbf{0.81$\pm$0.05}&447$\pm$18&&\textbf{0.73$\pm$0.07}&188$\pm$9&\textbf{0.61$\pm$0.05}&292$\pm$6&\textbf{0.46$\pm$0.05}&496$\pm$4&\\
        &NN--FC1&&0.71$\pm$0.14&2&0.75$\pm$0.06&3&0.49$\pm$0.12&5&&0.60$\pm$0.08&2&0.45$\pm$0.05&3&0.29$\pm$0.05&5&\\ \cmidrule{1-2} \cmidrule{4-9} \cmidrule{11-17}
        &kNN&&0.89$\pm$0.06&200&0.82$\pm$0.06&300&0.74$\pm$0.07&500&&0.66$\pm$0.07&200&0.49$\pm$0.07&300&0.33$\pm$0.06&500&\\
        &kNN + C&&0.88$\pm$0.06&\textbf{64$\pm$20}&0.81$\pm$0.04&\textbf{128$\pm$22}&0.73$\pm$0.06&\textbf{255$\pm$33}&&0.66$\pm$0.07&\textbf{119$\pm$18}&0.51$\pm$0.06&\textbf{214$\pm$19}&0.35$\pm$0.04&\textbf{414$\pm$16}&\\ \bottomrule[1.25pt]
    \end{tabular}
    \label{table:condensing_impact}
\end{table*}

The proposed solutions have been validated on two different application scenarios (described in Section~\ref{subsct:datasets}), two types of concept drift (defined in Section~\ref{subsct:changes}) and three different Micro-Controller Units (MCUs) from STMicroelectronics (whose technological details are given in Section~\ref{subsct:stm_porting}). In addition, the rest of the Section is organized as follows. Section~\ref{subsct:settings} discusses the experimental settings, whereas Sections~\ref{subsct:condensing_impact} and~\ref{subsct:algorithms_analysis} provides the experimental results. Finally, Section~\ref{subsct:stm_porting} presents the porting of the Hybrid Tiny $k$NN algorithm on the three considered MCUs.

It is crucial to point out that TML in presence of concept drift is a completely new research area and, to the best of our knowledge, this is the first work in the related literature proposing adaptive mechanisms for TML running on MCUs.
\subsection{Application Scenarios and Datasets}
\label{subsct:datasets}
In the experimental section, the following two application scenarios have been considered:
\begin{itemize}
    \item the \emph{speech-command identification} scenario whose goal is to correctly recognize an user-speech command present in a one-second long audio clip. For this purpose, the \emph{Synthetic Speech Commands Dataset}~\cite{buchner2017synthetic,warden2018speech} has been considered. This dataset comprises 30 classes of commands, corresponding, for example, to ``\emph{up}'', ``\emph{left}'', ``\emph{yes}'', ``\emph{go}'', or a number from ``\emph{zero}'' to ``\emph{nine}''. Moreover, the audio files within the dataset comprise different kinds of voices as well as different types of noisy classes.
    \item the \emph{image classification} scenario whose goal is to classify an image containing exactly one object. The well-known ImageNet~\cite{deng2009imagenet} dataset, comprising 1000 classes, has been considered.
\end{itemize}
\subsection{The Considered Concept Drift affecting $\mathcal{P}$}
\label{subsct:changes}
Two different kinds of concept drift affecting the data-generation process $\mathcal{P}$ have been considered:
\begin{itemize}
    \item the addition of noise on $x$. This type of concept drift models the scenario where a failure on the microphone acquiring the audio clip occurs. To achieve this goal, the noisy variant of each class within the \emph{Synthetic Speech Commands Dataset} is considered after the change;
    \item a change in the classification problem, i.e., a variation in the set of classes $\Lambda$.
\end{itemize}
%
%
\subsection{Experimental Settings}
\label{subsct:settings}
In this experimental analysis, the considered feature extractor $\varrho$ refers to the first layer of the well-known ResNet-18 CNN~\cite{he2016deep}. This layer comprises a convolutional layer with 64 7x7 three-dimensional filters with stride 2, a batch-normalization layer, a ReLU non-linearity, and a 3x3 max-pooling layer with stride 2. The dimensionality reduction operator $\varsigma$ discards 63 out of 64 filters by keeping only the one with the highest mean activation on the ImageNet benchmark. Consequently, the resulting DL model $\varsigma \circ \varrho$ is a single 7x7x3 filter that has 147 parameters and occupies $588$B with a 32-bit floating-point representation.

In the \emph{speech-command identification} scenario, the audio waveform (sampled at $f_a=22050$ Hz) is converted into a spectrogram through a Short-Time Fourier Transform with windows of size $n_{fft}=512$ and a step $h_l=512$ and then converted into a colored one by means of a colormap. In the \emph{image classification}, instead, the images are resized to 224x224x3 before being passed as input to $\mathcal{D}$. The resulting one-second audio has a memory footprint of $88\,200$B, the image $602\,112$B, whereas the resulting colored spectrogram of size 257x44x3 requires $135\,696$B.

The change always occurs after half of the available data, i.e., 500 samples per class in the \emph{image classification scenario} and 750 in the \emph{speech command identification} one. Finally, $100$ samples per class are provided to all the algorithm as initial training set $\mathcal{T}$, i.e., $\left|\mathcal{T}\right|=100\cdot\lambda$.
Experimental results are averaged over twenty runs. 
\subsection{Evaluating Effects of the Pre-Processing Through Condensing}
\label{subsct:condensing_impact}
\input{figure_algorithm2.tex}
The first aspect considered in this experimental analysis aims at studying the impact of condensing the kNN $\mathcal{K}$ training set $\mathcal{T}$ with Algorithm~\ref{algorithm:condensed_nearest_neighbor}. To achieve this goal, the proposed TML-CD $\mathcal{D}$ is configured with the block $\varsigma \circ \varrho$ presented in Section~\ref{subsct:settings} and an initial training set $\mathcal{T}$ with size $\left|\mathcal{T}\right|=100\cdot\lambda$. Then, the classification capabilities of $\mathcal{D}$ when condensing is employed are evaluated in stationary conditions, i.e., no adaptation is carried out during the operational life of $\mathcal{D}$. In addition to $\mathcal{D}$ without or with the initial condensing (referred to as $k$NN and $k$NN+C in the following, respectively), the comparison comprises two well-known classifiers, i.e., a Support Vector Machine (SVM) and a single fully-connected layer neural network classifier (NN--FC1). 
Both the classifiers are applied on the same features of the kNN $\mathcal{K}$, i.e., those extracted by $\varsigma \circ \varrho$ (on the initial training set $\mathcal{T}$). Moreover, the SVM is trained until convergence, whereas the NN--FC1 is trained for 3 epochs with stochastic gradient descent, no momentum, and a learning rate $\eta \in \left[1e^{-2}, 5e^{-3}, 1e^{-3}, \ldots, 1e^{-5}\right]$. In our experiments, only the best performing NN--FC1 classifier is shown. We emphasize that both the SVM and the NN are characterized by an unfeasible training procedure in MCUs, so they cannot be considered for the on-device training phase.

Table~\ref{table:condensing_impact} shows the result in the proposed application scenarios with a different number of classes. The SVM and the NN--FC1 classifiers present the highest and worst accuracy in all the considered application scenarios, respectively. The proposed TML solution $\mathcal{D}$ (without condensing) shows accuracies smaller than the SVM classifier by 4 to 8\% in \emph{speech command identification} and 7 to 13\% in \emph{image classification} scenarios. As expected, condensing the training set $\mathcal{T}$ has a limited impact on the accuracy (at most 1\% drop in \emph{speech command identification} scenario), but it allows to reduce the memory requirements significantly. Without considering the NN classifier, $\mathcal{D}$ with condensing is the algorithm providing the lowest memory demand.\footnote{The NN--FC1 memory footprint corresponds to that of its weights, which are equal to the number of classes multiplied by the size of classifier inputs. Since the input size is the same for all the classifiers, the NN-FC1 memory is that of $\lambda$ samples.} In the \emph{speech command identification} scenario, the number of stored samples indeed ranges from 32 to 50\% of the provided samples (with $\lambda$ from 2 to 5), representing the 46 to 57\% of the ones required by the SVM, i.e., its support vectors. In the \emph{image classification} scenario, the memory saving is significantly lower, with the SVM retaining almost all the samples as support vectors and the condensed $\mathcal{D}$ storing 60 to 82\% of them.
From now on, the proposed TML-CD $\mathcal{D}$ is assumed to always rely on condensing algorithm in the configuration and testing stages (when available).
\subsection{Experimental Results in Presence of Concept Drift}
\label{subsct:algorithms_analysis}
Figure~\ref{fig:results_algorithms} compares the three proposed adaptive algorithms, i.e., CIT, Active Tiny $k$NN, and Hybrid Tiny $k$NN, in the two considered scenarios with different numbers of classes $\lambda$. We considered two different figures of merit: the mean $\pm$ std \emph{accuracy} (the curve of each experiment is the convolution of the correct predictions of each experiment with a 100-dimensional filter with all values of $0.01$) and \emph{memory footprint}, measured as the number of samples within the training set $\mathcal{T}$, over all the experiments. It is crucial to point out that the memory footprint does not include any other auxiliary source of memory, e.g., the history window $W$ of the Active Tiny-$k$NN algorithm (that has a size of $\varpi=100\cdot\lambda$).

As a comparison, this experimental analysis includes also a continuously learning single-layer fully-connected classifier (NN--FC1) operating on the features extracted by $\varsigma \circ \varrho$ and performing a back-propagation step for each incoming sample. The NN--FC1 is trained for 3 epochs with stochastic gradient descent, no momentum, and a learning rate $\eta$, with $\eta \in \left[1e^{-2}, 5e^{-3}, 1e^{-3}, \ldots, 1e^{-5}\right]$. Moreover, during the testing stage, the learning rate for back-propagation might be reduced ($\eta \in \left[1e^{-3}, 5e^{-4}, 1e^{-4}, \ldots, 1e^{-6}\right]$). Among all the possible combinations, Figure~\ref{fig:results_approximated} shows only the one with the largest accuracy. 

In more detail, Figure~\ref{fig:results_algorithms} shows the accuracy and memory footprint in five different configurations: the \emph{Speech Command Identification scenario} where one-class changes with $\lambda=\{2,3,5\}$  classes (Figures~\ref{fig:results_audio_change_class},~\ref{fig:results_audio_change_class_3}, and~\ref{fig:results_audio_change_class_5}) and with the introduction of noise with $\lambda=2$ (Figure~\ref{fig:results_audio_noise}), and the \emph{Image Classification} scenario where one-class changes with $\lambda=2$ (Figure~\ref{fig:results_image_change_class}).

The NN-FC1 baseline is the worst algorithm in almost all the cases, with low capabilities of recovering after change when $\lambda > 2$ (Figures~\ref{fig:results_audio_change_class_3} and~\ref{fig:results_audio_change_class_5}). The noise has a limited impact on the accuracy, as shown in Figure~\ref{fig:results_audio_noise}, whose small degradation is detected neither by the Active nor by the Hybrid Tiny $k$NN algorithms. With the other changes, the proposed algorithms work as expected. On the one hand, the (passive) CIT algorithm continuously improves over time, at the expense of an unbounded memory growth (in these experiments, none of the approaches detailed in Section~\ref{sct:condensing_in_time} to control it has been considered). Moreover, in all the considered scenarios, the slope of the samples' curve increases at the change time, highlighting the accuracy drop due to the change itself. On the other hand, the Active $k$NN algorithm is able to recover after a change keeping its memory footprint nearly constant and significantly lower than the size of history window $W$ (not shown in the Figure~\ref{fig:results_algorithms}) due to condensing.
Finally, the Hybrid Tiny $k$NN algorithm combines the advantages of both the CIT and the Active Tiny $k$NN. It can recover faster than the two other algorithms in all the considered scenarios and keep the memory footprint under control (by taking into account also the Active Tiny $k$NN history window memory footprint, the Hybrid Tiny $k$NN has the lowest footprint). Moreover, it shows the best accuracy, being overcome by CIT only when it saturates the maximum size of its training set $\mathcal{T}$ that is here fixed to $\varpi=100\cdot\lambda$. This effect is visible in particular in Figures~\ref{fig:results_audio_change_class_3} and~\ref{fig:results_audio_change_class_5}.
\subsection{Porting the Hybrid Tiny $k$NN on the STM32 MCUs}
\label{subsct:stm_porting}
\input{figure_approximation.tex}
\begin{table}[t]
    \centering
    \scriptsize
    \caption{The detailed memory footprint (with a 32-bit data type) of the Hybrid Tiny $k$NN on the STM32 MCUs. The size of the training set $\mathcal{T}$ includes those of samples (see $\varsigma \circ \varrho$ output), their labels (32bit), and their timestamps (32bit).}
    \subfloat[STM32H743 and STM32F767.]{
        \begin{tabular}{C{3.8cm}cC{1.7cm}C{1.7cm}@{}m{0pt}@{}}
                &&Size&Memory Footprint (B)&\\ \midrule[1.25pt]
                Audio ($t_a=1$ s, $f_a=22050$ Hz)&&1x22050&88\,200&\\
                Spectrogram ($n_{fft}= h_l=512$)&&257x44x3&135\,696& \\ \cmidrule{1-1}\cmidrule{3-4}
                $\varsigma \circ \varrho$ (1 convolutional filter 7x7x3)&&7x7x3&588&\\
                $\varsigma \circ \varrho$ output&&65x11&2\,860&\\ \cmidrule{1-1}\cmidrule{3-4}
                $\mathcal{K}$'s Training Set $\mathcal{T}$&&50&143\,400&\\ \midrule
                Total&&&370\,744& \\ \bottomrule[1.25pt]
        \end{tabular} 
        \label{tab:solutionSTM32_h7_f7}   
    }
    \subfloat[STM32F401.]{
        \begin{tabular}{C{3.8cm}cC{1.7cm}C{1.7cm}@{}m{0pt}@{}}
            &&Size&Memory Footprint (B)&\\ \midrule[1.25pt]
            Audio ($t_a=1$ s, $f_a=4410$ Hz)&&1x4410&17\,640&\\
            Spectrogram ($n_{fft}= h_l=128$)&&64x35x3&27\,300& \\ \cmidrule{1-1}\cmidrule{3-4}
            $\varsigma \circ \varrho$ (1 convolutional filter 7x7x3)&&7x7x3&588&\\
            $\varsigma \circ \varrho$ output&&17x9&612&\\ \cmidrule{1-1}\cmidrule{3-4}
            $\mathcal{K}$'s Training Set $\mathcal{T}$&&50&31\,000&\\ \midrule
            Total&&&77\,140& \\ \bottomrule[1.25pt]
        \end{tabular} 
        \label{tab:solutionSTM32_f4}   
    }
    \label{tab:solutionSTM32}
\end{table}
\begin{table}[!t]
    \centering
    \scriptsize
    \caption{The experimental execution times, measured in milliseconds, on the STM32 MCUs. The measured times are: the spectrogram processing $t_p$, the feature extraction $t_{\varsigma\circ\varrho}$, and the $\mathcal{K}$ prediction with 10 and 50 samples within $\mathcal{T}$. The time $t_{\mathcal{K},50}$ also shows the worst prediction+adaptation time.}
    \begin{tabular}{C{2.5cm}cC{1cm}C{1cm}C{1cm}C{1cm}@{}m{0pt}@{}}
        MCU&&$t_p$&$t_{\varsigma\circ\varrho}$&$t_{\mathcal{K},10}$&$t_{\mathcal{K},50}$& \\ \midrule[1.25pt]
        STM32H743ZI&&22.9&18.6&2.0&2.9--6.2& \\ \cmidrule{1-1}\cmidrule{3-7}
        STM32F767ZI&&53.8&41.5&2.2&4.0--12.1&\\ \cmidrule{1-1}\cmidrule{3-7}
        STM32F401RE&&34.6&43.1&2.2&4.6--136.0&\\ \bottomrule[1.25pt]
    \end{tabular}
    \label{tab:stm32_time}
\end{table}
The aim of this section is to show the technological feasibility of the proposed Hybrid Tiny $k$NN algorithm in the \emph{Speech command identification} scenario. To achieve this goal, we considered the following three different MCUs: 
\begin{itemize}
    \item The \emph{STM32H743} is a high-performance MCU having a 480 MHz Cortex-M7 processor, 1024 KB of RAM (split into five blocks of different speed), and 2048 KB of Flash memory;
    \item The \emph{STM32F767} is a high-performance MCU having a 216 MHz Cortex-M7 processor, 512 KB of RAM, and 2048 KB of Flash memory;
    \item The \emph{STM32F401} is a general-purpose MCU having a 84 MHz Cortex-M4 processor, 96 KB of RAM, and 512 KB of Flash memory.
\end{itemize}
The main technological constraint imposed by such board is the one on the memory, i.e., the maximum memory footprint of the Hybrid $k$NN algorithm cannot overcome the available RAM of each MCU (in the case of \emph{STM32H743} that limit is lowered to 512KB, i.e., the size of the fastest RAM block). To satisfy this memory constraint, we set the maximum training set size of the Hybrid Tiny $k$NN algorithm to $\left|\mathcal{T}\right|=50$. In addition, for the \emph{STM32F401} board, the sampling frequency $f_a$ is reduced from $f_a=22050$Hz to $f_a=4410$Hz as suggested in~\cite{disabato2021birdsong}. This guarantees a strong reduction in the memory footprint of the input audio ($17\,640$B), the generated spectrogram with windows of size $n_{fft}=128$ and step $h_l=128$ (65x35x3 occupying $27\,300$B), and on the output of the feature extractor (17x9 occupying $588$B).
Table~\ref{tab:solutionSTM32} details the memory footprint of the Hybrid Tiny $k$NN deployed for the STM32H743 and STM32F767 (Table~\ref{tab:solutionSTM32_h7_f7}) and the STM32F401 (Table~\ref{tab:solutionSTM32_f4}).

Figure~\ref{fig:results_approximated} shows the effects of such technological choices in the considered scenario when a class change after half samples with $\lambda=\{2,3\}$. In that figure, the baseline is the Hybrid Tiny $k$NN algorithm introduced in Section~\ref{subsct:algorithms_analysis}, where $\varpi=100\cdot\lambda$. The algorithms deployed on the MCUs exhibit a constant accuracy until the change with a minimum gain due to passive adaptation (limited by the constraint on the training set size). The gap w.r.t. the baseline algorithm is between 4 and 10\% (15\% to 25\% when varying also the acquisition frequency). After the change, the algorithm with $\left|\mathcal{T}\right|=50$ recovers as fast as the baseline, then the effects of the passive updates create a gap in the accuracy. The algorithm that considers a variation also in the sampling frequency, instead, shows a larger drop in accuracy (about 10 to 15\%), but its passive updates are able to recover after the change, slightly improving the accuracy over time.

Table~\ref{tab:stm32_time} reports the experimental execution times of the $\mathcal{D}$ blocks in the considered MCUs. More in detail, the measured quantities are: the processing time $t_p$ needed to transform the acquired 1s audio into a spectrogram, the feature extractor and dimensionality reduction blocks' execution $t_{\varsigma\circ\varrho}$, and the $k$NN $\mathcal{K}$ prediction time $t_{\mathcal{K}, \left|\mathcal{T}\right|}$ with two different cardinalities of its training set, 10 and 50 (the latter also shows the worst measured prediction time when an adaptation has been made). Results are particularly interesting. In particular, the processing and the feature extraction are the two predominant times, requiring $41.5$ms and $95.3$ms on a high-performance MCUs (the STM32H7 and the STM32F7) and $77.7$ms on a general-purpose one (the STM32F4, although on a smaller spectrogram). The $\mathcal{K}$'s prediction and the adaptation, when employed, are negligible w.r.t. the other times, being the $7$ and the $15$\% on the STM32H7, the $4$\%  and the $13$\% on the STM32F7, and the $6$ and the $175$\% on the STM32F4 (whose adaptation is the only exception). As a final remark, the total time required from the processing of the acquired audio to the final prediction, including the possible adaptations, is significantly lower than that of the acquisition, showing the effectiveness of the proposed Hybrid Tiny $k$NN algorithm on three real MCUs.
    %
    \section{Conclusion}
    \label{sct:conclusions}
   For the first time in the literature, this paper introduced an adaptive Tiny Machine Learning solution for Concept Drift. This solution, characterized by a hybrid approach integrating an active and a passive adaptation step, takes into account the technological constraints on memory, computation, and energy typically characterizing embedded systems and IoT units it runs on. The proposed solution has been deployed to three different micro-controller units with 96 to 512KB of RAM, showing its feasibility in real-world scenarios and on off-the-shelf technological units.
   
   Future work will encompass the definition of advanced memory control mechanisms on passive updates (e.g., by deepening the suggested probabilistic approach), further optimization of the $k$NN memory requirements, the definition of learning mechanisms at the feature extractor block, and the exploration of sparse or quantized solutions for the TML algorithms.

    
    %

    %

    %
    %

    \ifCLASSOPTIONcaptionsoff
    \newpage
    \fi

    
    
    %
    \bibliographystyle{IEEEtran}
    \bibliography{bibliography.bib}
    
    %
    
    %
    %
    %
    
    
    

\end{document}